\documentclass[runningheads]{llncs}
\usepackage[T1]{fontenc}
\usepackage{graphicx}
\usepackage{biblatex}
\usepackage{multirow}
\usepackage{amssymb,amsfonts} 
\usepackage{amsmath}
\usepackage{float}
\usepackage{pdfcomment}
\usepackage{hyperref}
\hypersetup{
    colorlinks=true,
    linkcolor=blue,
    filecolor=magenta,      
    urlcolor=cyan,
    pdftitle={Overleaf Example},
    pdfpagemode=FullScreen,
    }

\begin{document}
\title{Enhanced Deep Learning DeepFake Detection Integrating Handcrafted Features}
\titlerunning{Enhanced Deep Learning DeepFake Detection}

\author{Alejandro Hinke-Navarro\inst{1} \and
Mario Nieto-Hidalgo\inst{1}\orcidID{0000-0003-0623-6455} \and
Juan M. Espín\inst{1}\orcidID{0000-0001-6521-7890} \and
Juan E. Tapia\inst{2}\orcidID{0000-0001-9159-4075}
}%
\authorrunning{A. Hinke et al.}

\institute{Facephi Biometrics S.A., Alicante, Spain.\\
\email{\{alejandrohinke, marionieto, jmespin\}@facephi.com}\\
\and
da/sec-Biometrics and Internet Security Research Group, Darmtstadt, Germany\\
\email{juan.tapia-farias@h-da.de}
}
\maketitle              
\begin{abstract}
The rapid advancement of deepfake and face swap technologies has raised significant concerns in digital security, particularly in identity verification and onboarding processes. Conventional detection methods often struggle to generalize against sophisticated facial manipulations. This study proposes an enhanced deep‑learning detection framework that combines handcrafted frequency‑domain features with conventional RGB inputs. This hybrid approach exploits frequency and spatial domain artifacts introduced during image manipulation, providing richer and more discriminative information to the classifier. Several frequency handcrafted features were evaluated, including the Steganalysis Rich Model, Discrete Cosine Transform, Error Level Analysis, Singular Value Decomposition, and Discrete Fourier Transform.

\end{abstract}

\section{Introduction}\label{sec:introduction}
Image manipulation has become a widely discussed topic over the years, with its detection posing an increasingly complex challenge because of the rapid advancements in generative techniques. Among the various forms of digital face manipulation, key methods include face swap, identity swap, attribute manipulation, and entire face synthesis \cite{tolosana2020deepfakes}.

Face swapping involves replacing a target individual's face with another person's, effectively altering the subject's appearance while retaining their original context. In contrast, full-face synthesis refers to the complete generation of facial images from scratch using advanced generative models, such as Generative Adversarial Networks (GANs) or diffusion models. These techniques enable the creation of highly realistic facial representations, often indistinguishable from authentic images.

Nowadays, most users perceive this technology as harmless entertainment; however, it is increasingly being misused for malicious purposes, such as spreading fake news, generating illicit content, and engaging in political manipulation, among others. These harmful applications have a significant impact on social media, undermining trust and contributing to a crisis of authenticity in digital content across the Internet.

Traditional methods, based on RGB pixel values and convolutional neural networks (CNNs), perform well on intra-datasets but struggle to generalize across unseen datasets \cite{tolosana2020deepfakes}. This limitation arises because artifacts indicative of manipulation can be significantly diminished due to factors such as image compression or manual editing, making it challenging for these models to detect subtle inconsistencies across diverse sources.

To address this challenge, the following research questions are explored:
\begin{itemize}
    \item How can generalization across datasets be improved?
    \item Which frequency-domain representations are most effective?
    \item How can frequency-domain features be integrated into deep learning models?
\end{itemize}

This work focuses specifically on identity swapping and full-face synthesis, two of the most prevalent manipulation techniques in digital media.

The main contributions of this work are:
\begin{itemize}
    \item A study of frequency-domain features for face manipulation detection.
    \item An evaluation of several handcrafted features, identifying Discrete Cosine Transform as the most effective.
    \item A demonstration that minimum score-level fusion between intensity pixel values and frequency features yields improved performance over baseline models.
\end{itemize}

The remainder of this article is structured as follows: \textbf{Section} \ref{Related Works} reviews related work on deepfake and face manipulation detection. \textbf{Section} \ref{Methods} describes the datasets and the proposed method. \textbf{Section} \ref{Experiments} reports experimental results. Finally, \textbf{Section} \ref{Conclusions} concludes the paper and outlines directions for future research.

\section{Related Works} \label{Related Works}

Traditional methods based on intensity values (RGB) images and convolutional neural networks (CNNs) have demonstrated high performance on intra-datasets but lower generalization capabilities to perform with high rates on cross-datasets. To overcome this challenge, new approaches like frequency domain have been explored based on the changes in frequencies (high and low) that are produced when the image is manipulated. A similar effect is dedicated to compression.

Conventional deepfake detection approaches predominantly leverage spatial domain features extracted from pixel values across RGB images using deep neural networks. 

Studies such as Luo et al. \cite{luo2021generalizing} highlight that CNN-based models often overfit to method-specific color textures, which limits their generalization capabilities when tested against unseen manipulations.

Similarly, the work by Ibsen et al. \cite{dong2023implicit} emphasizes that RGB-only models struggle to differentiate real from synthetic faces when exposed to novel generative models or post-processing operations.
These limitations underline the need for more robust detection techniques that incorporate additional feature representations beyond RGB data.

Recent advances in deepfake detection have highlighted the effectiveness of frequency-domain analysis in identifying manipulated content \cite{wang2023frequency, luo2021generalizing}. 

Wang et al. \cite{wang2023frequency} introduced a Frequency Domain Filtered Residual Network, which enhances detection robustness by fusing wavelet-transformed frequency information with RGB data, particularly improving performance on compressed deepfake images. 

Luo et al. \cite{luo2021generalizing} showed that multi-scale SRM filtering strengthens cross-dataset generalization by detecting high-frequency noise residuals. 

More recently, Tan et al. \cite{tan2024frequency} proposed FreqNet, a frequency-aware model that enhances deepfake detector generalization by learning high-frequency features independently of their source. 

Li et al. \cite{li2024freqblender} introduced FreqBlender, a method that synthesizes pseudo-fake faces by manipulating frequency information, improving the learning of generic forgery traces, and enhancing detection accuracy.

Tapia et al. \cite{tapia2023face} also demonstrate that frequency-based filters can be used to detect digital manipulation attacks, such as Morphing.

Rahaman et al. \cite{rahaman2019spectral} introduced the concept of spectral bias, demonstrating through Fourier analysis that neural networks exhibit a learning preference for low-frequency functions. This spectral bias explains why neural networks often generalize well to natural data and highlights the robustness of low-frequency components to parameter perturbations.

Many of these findings suggest that the frequency domain contains valuable information that can be effectively leveraged to improve the detection of manipulated images.

\vspace{-0.3cm}
\section{Proposed Method} 
\begin{figure}[H]
    \centering
    \pdftooltip{\includegraphics[width=12cm]{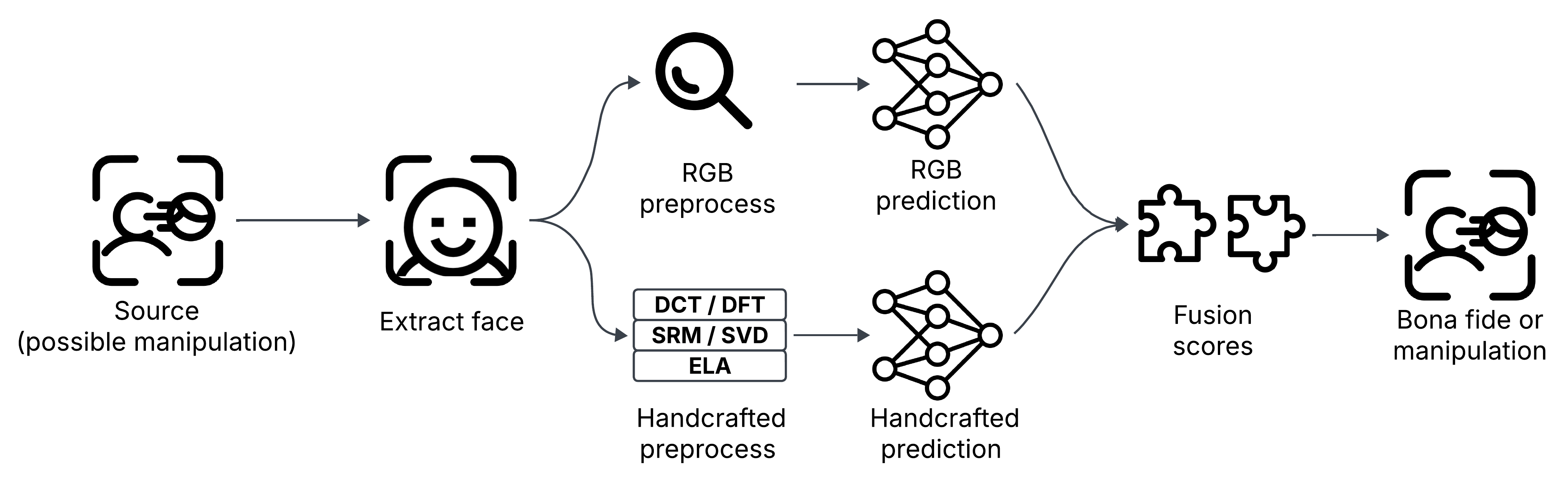}}{Diagram showing a pipeline for detecting image manipulation. The process begins with a source image (which may be manipulated), followed by face extraction. The extracted face is processed in two parallel paths: one for RGB preprocessing and another for handcrafted preprocessing (using DCT, DFT, SRM, SVD, or ELA). Each path feeds into its own neural network: RGB prediction and handcrafted prediction. The outputs are fused to produce a final score, which classifies the image as either bona fide or manipulated}
    \caption{Manipulation Attack Detection Framework.}
    \label{fig:flow}
\end{figure}

This work proposes a method for detecting face digital manipulation attacks, based on handcrafted frequency features and fusion with intensity values at the score level. Several datasets were employed to evaluate the generalization capabilities of the proposed approach. A diagram illustrating the method is presented in Figure \ref{fig:flow}.

\vspace{-0.3cm}

\label{Methods}
\subsection{Datasets}
\vspace{-0.3cm}
In this study, six different datasets of digitally manipulated face images were used:

\begin{itemize}
    \item \textbf{FaceForensics++\cite{rossler2019faceforensics++}:} A widely used dataset for deepfake detection, comprising 4,320 videos, including 720 original videos sourced from YouTube and 3,600 manipulated videos generated using FaceShifter, FaceSwap, Face2Face, Deepfakes, and NeuralTextures. The official dataset split was followed, with 720 videos for training, 140 for validation, and 140 for testing.     
    Five random frames per video were used in this study.

    \item \textbf{Celeb-DF\cite{li2020celeb}:} A deepfake dataset specifically designed for identity-swapping manipulations, containing 5,639 deepfake videos generated from 590 original videos sourced from YouTube. Due to its real-world origin, the dataset is highly compressed, often exhibiting lower visual quality and compression artifacts, making detection more challenging. 
    Only one frame per video was used in this study.

    \item \textbf{DeepfakeTIMIT\cite{korshunov2018deepfakes}:} This dataset comprises videos where faces are swapped using a GAN-based approach developed from the original autoencoder-based Deepfake algorithm. It includes 620 videos with faces swapped, using the VidTIMIT database as the source. Two different qualities are provided: lower quality (LQ) with $64\times64$ input/output size models and higher quality (HQ) with $128\times128$ size models. 
    One frame per video was used in this study.

    \item \textbf{DeePhy\cite{narayan2022deephy}:} This dataset employs sequential face swapping based on a phylogenetic approach. It contains 468 spoof videos sourced from YouTube, encoded in MPEG4 format with a resolution of 720p, using a single frame per video.
    One frame per video was used in this study.
    
    \item \textbf{Defacto\cite{mahfoudi2019defacto}:} This dataset includes face-swapped images generated from MS-COCO images through automated forgery generation techniques, resulting in semantically meaningful and detailed manipulations. It contains 3,000 spoof images of variable sizes.
    One frame per video was used in this study.
    
    \item \textbf{SWAN-DF\cite{Korshunov_IJCB_2023}:} The first high-fidelity publicly available dataset of realistic audio-visual deepfakes, where both faces and voices appear and sound like the target person. Based on the public SWAN database of real videos recorded in HD on iPhone and iPad Pro, it includes 30 pairs of manually selected individuals. Faces and voices were swapped using several autoencoder-based face-swapping models and blending techniques from DeepFaceLab, along with voice conversion methods such as YourTTS, DiffVC, HiFiVC, and FreeVC.
    A random selection of 10\% of the dataset was used in this study.
\end{itemize}
Table \ref{tab:datasets_summary} shows a summary of all the datasets used in this research.

\begin{table}[H]
\scriptsize
\caption{Summary of datasets. DF, FS, NT, FSW, and F2F represent DeepFake, FaceShifter, NeuralTransfer, FaceSwap, and Face-to-Face, respectively.}
\label{tab:datasets_summary}
\centering
\begin{tabular}{|l|l|l|}
\hline
\textbf{Database} & \textbf{Nº of Images}  & \textbf{Manipulation algorithm} \\ \hline
FF++             & 25,000 fake, 5000 real & DF, FS, NT, FSW, F2F                 \\ \hline
CelebDF          & 5,639 fake, 590 real  & Improved DF                     \\ \hline
DeepfakeTIMIT    & 640 fake             & GAN-based (face swap-GAN)        \\ \hline
DeePhy           & 468 fake, 100 real   & Phylogenetic sequential FS      \\ \hline
Defacto          & 3,000 fake, 200 real  & Automated semantic FS           \\ \hline
SWAN-DF          & 11,940 fake           & Autoencoder-based (DeepFaceLab) \\ \hline
\end{tabular}
\end{table}

\subsection{Metrics}
To evaluate the effectiveness of the proposed method, the ISO/IEC 30107-3 was followed\footnote{\url{https://www.iso.org/standard/79520.html}}. Detection Equal Error Rate (D-EER) metric was employed, which represents the point at which the Attack Presentation Classification Error Rate (APCER) and the Bona fide Presentation Classification Error Rate (BPCER) are equal. The APCER indicates the proportion of attack presentations incorrectly classified as bona fide (false positives), while BPCER denotes the proportion of bona fide presentations incorrectly classified as attacks (false negatives). 
A lower D-EER value reflects the higher accuracy and robustness of the detection system. D-EER is widely used in biometric systems and forgery detection tasks due to its balanced assessment of both types of error rates, providing a comprehensive measure of system performance.

\subsection{Feature Extraction}
Several feature extraction techniques based on handcrafted features have been employed to distinguish between bona fide and digitally manipulated images \cite{tapia2023face}. 
In this study, five frequency handcrafted feature extraction methods were used individually and in combination to improve the detection of manipulated faces: Color (RGB), which is represented by the pixel values, Discrete Cosine Transform (DCT), Steganalysis Rich Model (SRM), Discrete Fourier Transform (DFT), Error Level Analysis (ELA), and Singular Value Decomposition (SVD). All features were extracted from grayscale versions of the images using to emphasize structural and frequency domain characteristics, except for color pixel values of RGB images. 
\vspace{-0.3cm}

\subsubsection{Discrete Cosine Transform (DCT)}
DCT is a widely used technique in image processing that transforms spatial domain information into the frequency domain. It decomposes an image into a sum of cosine functions oscillating at different frequencies, which helps detect hidden artifacts introduced during manipulations, particularly in compressed images, as it is a core component of popular formats like JPEG that exploit frequency information for efficient compression.

In this study, DCT was applied to the entire image as well as to sub-blocks of varying sizes, specifically $8\times8$, $12\times12$, $16\times16$, $20\times20$, and $24\times24$ pixels, with the $20\times20$ configuration proving to be the most effective.

\subsubsection{Steganalysis Rich Model (SRM)}

SRM is a feature extraction technique commonly used in digital forensics to detect hidden modifications in images. It focuses on capturing high-frequency noise patterns that arise from manipulation processes. In this study, an SRM filter using a kernel described in Eq. \ref{fig:srm_filter} was applied to the grayscale images to enhance edge detection and expose subtle alterations.

\begin{center} 
\begin{equation}
\label{fig:srm_filter}
SRM=\left\lceil
\begin{matrix}
0.0 & 1.0 & 0.0 \\
1.0 & -4.0 & 1.0 \\
0.0 & 1.0 & 0.0
\end{matrix}
\right\rceil 
\end{equation} 
\end{center}

This filter emphasizes discrepancies in the high-frequency domain by highlighting regions where pixel intensities exhibit irregular patterns, which are often indicative of tampering.

\subsubsection{The Discrete Fourier Transform (DFT)}
DFT converts an image from the spatial domain to the frequency domain, representing it in terms of sinusoidal components. This transformation helps analyze periodic patterns and identify inconsistencies introduced by generative models or post-processing operations. It is particularly useful for detecting manipulation artifacts that manifest as unnatural frequency distributions.
\vspace{-0.3cm}

\subsubsection{Error Level Analysis (ELA)}
ELA is a forensic technique used to detect areas of an image that have undergone different levels of compression. Repeated compression of an image and comparison with the original reveal discrepancies in compression artifacts, which can indicate tampered regions. In this study, it was applied to grayscale images to identify potential manipulation traces based on differences in compression levels across various regions of the image. Areas with significant discrepancies often correspond to edited portions, making ELA a valuable tool for forgery detection.
\vspace{-0.3cm}

\subsubsection{Singular Value Decomposition (SVD)}
\begin{figure}[H]
    \centering
    \pdftooltip{\includegraphics[width=7.5cm]{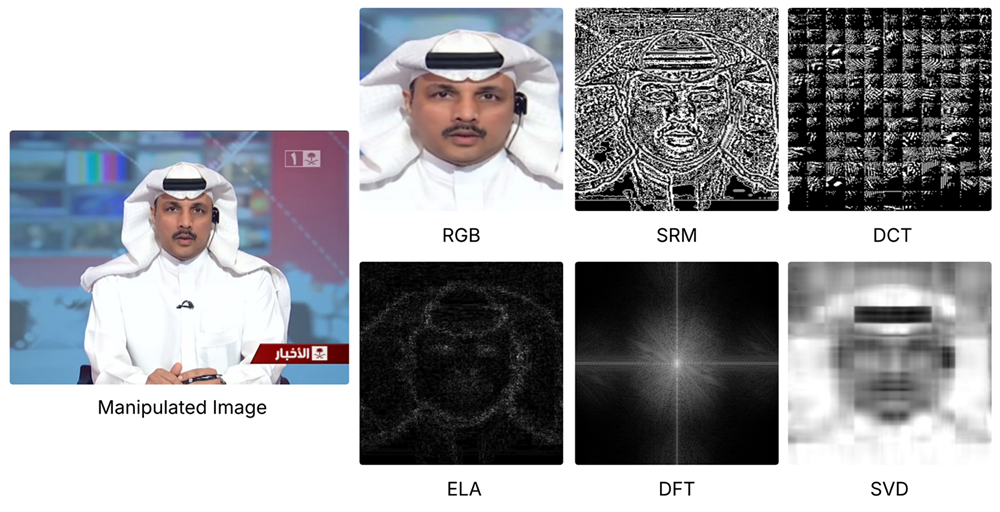}}{A manipulated news broadcast image is shown on the left. To the right, six derived feature representations of the face are displayed: RGB, SRM, DCT, ELA, DFT, and SVD. These highlight different frequency and structural artifacts used for manipulation detection.}
    \caption{Feature extraction example for a manipulated image.}
    \label{fig:feature_extraction}
\end{figure}

SVD is a matrix factorization technique that decomposes an image into three matrices, representing its intrinsic structure in terms of singular values and orthogonal components. It is effective in identifying structural changes caused by manipulations, as alterations typically disrupt an image's natural rank and singular value distribution. This study applied SVD to grayscale images to capture global structural inconsistencies with a component of 50. Figure \ref{fig:feature_extraction}, shows an example of the frequency features extracted.

\vspace{-0.4cm}

\subsection{Models}
\subsubsection{Preprocesing.}
The preprocessing pipeline begins by cropping faces and adding a 50\% padding around each crop. This wider margin exposes background context that face‑swap and similar attacks typically leave unaltered, allowing the network to contrast manipulated pixels with their unmodified surroundings. Subsequently, the images are resized to a fixed resolution of $384\times384$ pixels. 

The resized images serve as inputs to either EfficientNetV2 B0 \cite{tan2021efficientnetv2} or MobileViT-S \cite{mehta2021mobilevit} models, both of which are initialized using ImageNet pre-trained weights. EfficientNetV2-B0 was selected due to its well-balanced trade-off between computational efficiency and performance, making it suitable for deployment in resource-constrained environments. MobileViT-S, chosen for its compact size and rapid inference capability, leverages transformer-like attention mechanisms to capture detailed feature interactions through self-attention maps.

For handcrafted models, images are converted to grayscale. Several data augmentation techniques were employed during training to enhance model robustness, including horizontal flipping, random contrast adjustment, random brightness variation, random hue shifts, random saturation changes, and random JPEG compression, which were applied exclusively to manipulated images. This choice is motivated since GAN generated images often lack of compression artifacts, random JPEG compression was applied to manipulated samples to prevent the model from relying on this pattern and instead focus on manipulation-related traces.

Model weights were optimized using the AdaGrad algorithm with a minibatch size of $32$ and an initial learning rate of $1e-4$. Training was conducted for up to 225 steps, equivalent to approximately 65 epochs. For the MobileViT-S architecture, a patch size of $2$ was explicitly adopted. All training was performed using an NVIDIA A100 GPU.
\vspace{-0.3cm}

\subsubsection{Fusion at Score Level}
The fusion of scores involves combining the outputs of different models (RGB and Frequency) based on specific aggregation rules. The fusion strategies considered in this experiment include weighted fusion, where models contribute based on assigned importance; minimum fusion, which selects the lowest score among the models; mean fusion, which computes the average score; and maximum fusion, which takes the highest score from each model.

\section{Experiments} \label{Experiments}
Three experiments were proposed to show and compare the results with different frequency filters.

\subsection{Experiment 1: Handcrafted features benchmark}

All filters were trained and evaluated using the datasets mentioned in Table \ref{tab:datasets_summary} to measure the impact of each one.
\vspace{-0.3cm}

\begin{table}[H]
\centering
\scriptsize
\caption{D-EER \% for the different handcrafted features.}
\label{original_bonafides}

\setlength{\tabcolsep}{5pt} 
\renewcommand{\arraystretch}{1.2} 

\begin{tabular}{|cl|l|lll|}
\hline
\multicolumn{2}{|c|}{\multirow{2}{*}{}} & \textbf{Intra} & \multicolumn{3}{c|}{\textbf{Cross}}                                  \\ \cline{3-6} 
\multicolumn{2}{|c|}{}                  & FF++           & \multicolumn{1}{l|}{Celeb-DF} & \multicolumn{1}{l|}{Dephy} & Defacto \\ \hline
\multicolumn{1}{|c|}{\multirow{6}{*}{Effv2b0}}     & RGB & 6.20 & \multicolumn{1}{l|}{35.87} & \multicolumn{1}{l|}{11.72} & 30.00 \\ \cline{2-6} 
\multicolumn{1}{|c|}{}       & SRM      & 5.75           & \multicolumn{1}{l|}{51.53}    & \multicolumn{1}{l|}{14.11} & 61.10   \\ \cline{2-6} 
\multicolumn{1}{|c|}{}       & DCT      & 3.18           & \multicolumn{1}{l|}{46.96}    & \multicolumn{1}{l|}{8.19}  & 48.37   \\ \cline{2-6} 
\multicolumn{1}{|c|}{}       & ELA      & 8.58           & \multicolumn{1}{l|}{45.96}    & \multicolumn{1}{l|}{11.61} & 38.92   \\ \cline{2-6} 
\multicolumn{1}{|c|}{}       & DFT      & 35.23          & \multicolumn{1}{l|}{46.54}    & \multicolumn{1}{l|}{36.30} & 48.00   \\ \cline{2-6} 
\multicolumn{1}{|c|}{}       & SVD      & 31.69          & \multicolumn{1}{l|}{44.77}    & \multicolumn{1}{l|}{17.63} & 33.07   \\ \hline
\multicolumn{1}{|c|}{\multirow{6}{*}{MobileViT-S}} & RGB & 1.37 & \multicolumn{1}{l|}{36.76} & \multicolumn{1}{l|}{7.05}  & 31.97 \\ \cline{2-6} 
\multicolumn{1}{|c|}{}       & SRM      & 13.93          & \multicolumn{1}{l|}{53.98}    & \multicolumn{1}{l|}{25.83} & 53.50   \\ \cline{2-6} 
\multicolumn{1}{|c|}{}       & DCT      & 5.03           & \multicolumn{1}{l|}{49.18}    & \multicolumn{1}{l|}{20.02} & 50.50   \\ \cline{2-6} 
\multicolumn{1}{|c|}{}       & ELA      & 13.59          & \multicolumn{1}{l|}{46.11}    & \multicolumn{1}{l|}{9.44}  & 33.50   \\ \cline{2-6} 
\multicolumn{1}{|c|}{}       & DFT      & 40.27          & \multicolumn{1}{l|}{48.67}    & \multicolumn{1}{l|}{45.85} & 48.03   \\ \cline{2-6} 
\multicolumn{1}{|c|}{}       & SVD      & 35.43          & \multicolumn{1}{l|}{41.56}    & \multicolumn{1}{l|}{20.02} & 32.50   \\ \hline
\end{tabular}
\end{table}

Observing the cross-dataset performance in Table \ref{original_bonafides}, the results are generally suboptimal. These findings suggest that mismatched bona fide distributions due to dataset-specific conditions such as varying image acquisition settings, compression methods, and quality negatively affect model generalization. Experiment 2 explores a potential solution by using a single, consistent source of bona fide images from FaceForensics++(FF++).

\subsection{Experiment 2: Handcrafted features benchmark using FF++ bona fides}
\begin{figure}[]
    \centering
    \pdftooltip{\includegraphics[width=12cm]{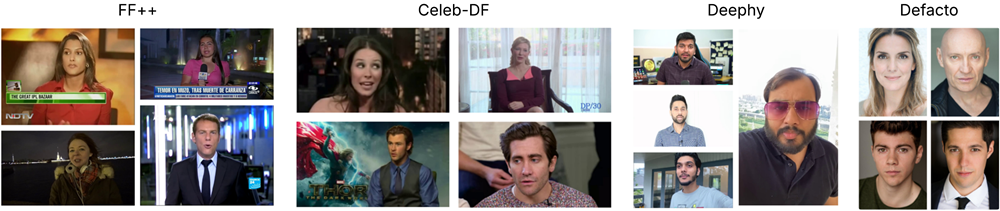}}{Grid of example bona fide face images from four deepfake-related datasets. From left to right: FF++ (four diverse broadcast-style video frames), Celeb-DF (four celebrity interview scenes), DeeperForensics (four webcam or selfie-style male faces), and Defacto (six high-quality portrait images of individuals with varied backgrounds and expressions). Each group illustrates the authentic image style typical of its respective dataset.}
    \caption{Bona fide samples distribution across datasets.}
 \label{fig:bonafide_distribution}
\end{figure}

Due to significant discrepancies in the distribution of bona fide images across the public datasets used in this study (see  Fig.~\ref{fig:bonafide_distribution}), only bona fide images from the FF++ dataset were employed for final model evaluation. This decision is supported by several considerations. 

\begin{itemize} \item \textbf{Heterogeneous capture conditions.} The datasets differ markedly in their image sources: some contain “in‑the‑wild” pictures taken under uncontrolled settings, whereas others include images acquired in controlled or studio environments. This mismatch produces considerable variation in visual characteristics and overall image quality.

\item \textbf{Compression and format inconsistencies.} Differences in compression schemes, file formats, and orientations introduce additional divergence, making cross‑dataset comparisons more difficult.

\item \textbf{Shortage of bona fide samples.} Several datasets provide only a small number of bona fide images, or none at all, which limits representativeness and reduces statistical reliability during evaluation.
\end{itemize}
Figure \ref{fig:bonafide_distribution} shows the different images from bona fide subsets.
\vspace{-0.3cm}

\vspace{-0.4cm}

\subsection{Experiment 3: Fusions at score level}
Building upon the findings from Experiment 2 in Table \ref{labelFF_bonafides}, RGB and DCT demonstrated superior performance. This experiment focuses on these two feature set. The DCT-based model exhibits strong detection capabilities for identity face swapping but performs poorly in full-face synthesis detection, whereas the RGB-based model shows the opposite trend.
The objective is to leverage the strengths of both spatial and frequency domains to enhance detection performance by exploring various fusion strategies.

\begin{table}[h]
\centering
\scriptsize
\caption{D-EER \% for the different handcrafted features using FF++ bona fides for every dataset. The highlighted numbers in bold indicate the best performance observed across the dataset.}
\label{labelFF_bonafides}
\begin{tabular}{|cc|c|ccccc|l|}
\hline
\multicolumn{2}{|c|}{\multirow{2}{*}{}} &
  \textbf{Intra} &
  \multicolumn{5}{c|}{\textbf{Cross}} &
  \multicolumn{1}{c|}{\multirow{2}{*}{\textbf{Avg}}} \\ \cline{3-8}
\multicolumn{2}{|c|}{} &
  FF++ &
  \multicolumn{1}{c|}{\begin{tabular}[c]{@{}c@{}}Celeb-\\ DF\end{tabular}} &
  \multicolumn{1}{c|}{\begin{tabular}[c]{@{}c@{}}Df-\\ Timit\end{tabular}} &
  \multicolumn{1}{c|}{Dephy} &
  \multicolumn{1}{c|}{Defacto} &
  \begin{tabular}[c]{@{}c@{}}Swan\\ DF\end{tabular} &
  \multicolumn{1}{c|}{} \\ \hline
\multicolumn{1}{|c|}{\multirow{6}{*}{Effv2b0}} &
  RGB &
  6.20 &
  \multicolumn{1}{c|}{17.45} &
  \multicolumn{1}{c|}{\textbf{6.39}} &
  \multicolumn{1}{c|}{9.43} &
  \multicolumn{1}{c|}{29.36} &
  15.58 &
  14.07 \\ \cline{2-9} 
\multicolumn{1}{|c|}{} &
  SRM &
  5.75 &
  \multicolumn{1}{c|}{17.12} &
  \multicolumn{1}{c|}{36.25} &
  \multicolumn{1}{c|}{22.95} &
  \multicolumn{1}{c|}{18.43} &
  24.17 &
  20.78 \\ \cline{2-9} 
\multicolumn{1}{|c|}{} &
  DCT &
  3.18 &
  \multicolumn{1}{c|}{\textbf{3.86}} &
  \multicolumn{1}{c|}{50.49} &
  \multicolumn{1}{c|}{12.38} &
  \multicolumn{1}{c|}{\textbf{9.93}} &
  29.35 &
  18.20 \\ \cline{2-9} 
\multicolumn{1}{|c|}{} &
  ELA &
  8.58 &
  \multicolumn{1}{c|}{8.22} &
  \multicolumn{1}{c|}{13.75} &
  \multicolumn{1}{c|}{14.57} &
  \multicolumn{1}{c|}{44.13} &
  33.21 &
  20.41 \\ \cline{2-9} 
\multicolumn{1}{|c|}{} &
  DFT &
  35.23 &
  \multicolumn{1}{c|}{24.00} &
  \multicolumn{1}{c|}{61.41} &
  \multicolumn{1}{c|}{53.14} &
  \multicolumn{1}{c|}{62.44} &
  69.29 &
  50.92 \\ \cline{2-9} 
\multicolumn{1}{|c|}{} &
  SVD &
  31.69 &
  \multicolumn{1}{c|}{35.61} &
  \multicolumn{1}{c|}{10.92} &
  \multicolumn{1}{c|}{25.53} &
  \multicolumn{1}{c|}{42.89} &
  36.04 &
  30.78 \\ \hline
\multicolumn{1}{|c|}{\multirow{6}{*}{\begin{tabular}[c]{@{}c@{}}Mobile\\ ViT-S\end{tabular}}} &
  RGB &
  \textbf{1.37} &
  \multicolumn{1}{c|}{\textbf{4.03}} &
  \multicolumn{1}{c|}{27.18} &
  \multicolumn{1}{c|}{\textbf{6.58}} &
  \multicolumn{1}{c|}{11.40} &
  \textbf{8.72} &
  9.21 \\ \cline{2-9} 
\multicolumn{1}{|c|}{} &
  SRM &
  13.93 &
  \multicolumn{1}{c|}{19.13} &
  \multicolumn{1}{c|}{36.73} &
  \multicolumn{1}{c|}{34.57} &
  \multicolumn{1}{c|}{19.11} &
  27.01 &
  25.91 \\ \cline{2-9} 
\multicolumn{1}{|c|}{} &
  DCT &
  \textbf{5.03} &
  \multicolumn{1}{c|}{7.40} &
  \multicolumn{1}{c|}{49.35} &
  \multicolumn{1}{c|}{25.14} &
  \multicolumn{1}{c|}{\textbf{9.53}} &
  34.27 &
  21.79 \\ \cline{2-9} 
\multicolumn{1}{|c|}{} &
  ELA &
  13.59 &
  \multicolumn{1}{c|}{10.40} &
  \multicolumn{1}{c|}{17.47} &
  \multicolumn{1}{c|}{17.81} &
  \multicolumn{1}{c|}{48.18} &
  46.31 &
  25.63 \\ \cline{2-9} 
\multicolumn{1}{|c|}{} &
  DFT &
  40.27 &
  \multicolumn{1}{c|}{36.06} &
  \multicolumn{1}{c|}{47.82} &
  \multicolumn{1}{c|}{52.66} &
  \multicolumn{1}{c|}{64.10} &
  66.48 &
  51.57 \\ \cline{2-9} 
\multicolumn{1}{|c|}{} &
  SVD &
  35.43 &
  \multicolumn{1}{c|}{34.87} &
  \multicolumn{1}{c|}{18.77} &
  \multicolumn{1}{c|}{25.72} &
  \multicolumn{1}{c|}{38.93} &
  41.95 &
  32.78 \\ \hline
\end{tabular}
\end{table}

It is essential to emphasise that model calibration prior to score fusion significantly impacts overall performance. Since FF++ served as the common source of bona fide images for each dataset, calibration was conducted by targeting a BPCER value. This approach ensures a consistent thresholding strategy across datasets.

In Table \ref{min_fusion}, the default protocol refers to the configuration obtained directly from training without applying any threshold calibration. For the RGB EfficientNet v2 b0 and DCT EfficientNet v2 b0 models, the "Default" model configuration yielded a BPCER of 19.27\% and 7.71\%, respectively.
In contrast, the RGB MobileViT-S and DCT MobileViT-S models achieved BPCERs of 4.02\% and 11.07\%, respectively, under the default conditions. 
"Protocol I" refers to each model being calibrated to BPCER 2.00\% before the fusion. "Protocol II" means that each model has been calibrated to BPCER 5.00\% before the fusion.

\begin{table}[H]
\caption{D-EER \% for the different fusions by a minimum score between RGB and DCT with the designed protocol. The highlighted numbers in bold indicate the best performance observed across the dataset.}
\label{min_fusion}
\centering
\scriptsize
\begin{tabular}{cl|l|lllll|}
\cline{3-8}
\multicolumn{2}{c|}{} &
  \textbf{Intra} &
  \multicolumn{5}{c|}{\textbf{Cross}} \\ \hline
\multicolumn{1}{|c|}{\textbf{Model}} &
  \textbf{Protocol} &
  FF++ &
  \multicolumn{1}{l|}{Celeb-DF} &
  \multicolumn{1}{l|}{Df.TIMIT} &
  \multicolumn{1}{l|}{Dephy} &
  \multicolumn{1}{l|}{Defacto} &
  SwanDF \\ \hline
\multicolumn{1}{|c|}{\multirow{3}{*}{\begin{tabular}[c]{@{}c@{}}RGB Effv2b0 +\\ DCT Effv2b0\end{tabular}}} &
  Default &
  2.01 &
  \multicolumn{1}{l|}{5.54} &
  \multicolumn{1}{l|}{\textbf{7.69}} &
  \multicolumn{1}{l|}{5.52} &
  \multicolumn{1}{l|}{13.28} &
  15.40 \\ \cline{2-8} 
\multicolumn{1}{|c|}{} &
  Protocol I &
  2.03 &
  \multicolumn{1}{l|}{4.73} &
  \multicolumn{1}{l|}{8.74} &
  \multicolumn{1}{l|}{5.52} &
  \multicolumn{1}{l|}{11.76} &
  15.97 \\ \cline{2-8} 
\multicolumn{1}{|c|}{} &
  Protocol II &
  1.99 &
  \multicolumn{1}{l|}{4.49} &
  \multicolumn{1}{l|}{9.22} &
  \multicolumn{1}{l|}{\textbf{5.33}} &
  \multicolumn{1}{l|}{11.44} &
  16.44 \\ \hline
\multicolumn{1}{|c|}{\multirow{3}{*}{\begin{tabular}[c]{@{}c@{}}RGB MobileViT-S +\\ DCT MobileViT-S\end{tabular}}} &
  Default &
  1.34 &
  \multicolumn{1}{l|}{2.98} &
  \multicolumn{1}{l|}{38.43} &
  \multicolumn{1}{l|}{8.57} &
  \multicolumn{1}{l|}{9.80} &
  11.74 \\ \cline{2-8} 
\multicolumn{1}{|c|}{} &
  Protocol I &
  1.34 &
  \multicolumn{1}{l|}{2.52} &
  \multicolumn{1}{l|}{34.55} &
  \multicolumn{1}{l|}{7.05} &
  \multicolumn{1}{l|}{8.23} &
  9.73 \\ \cline{2-8} 
\multicolumn{1}{|c|}{} &
  Protocol II &
  1.34 &
  \multicolumn{1}{l|}{2.98} &
  \multicolumn{1}{l|}{38.43} &
  \multicolumn{1}{l|}{8.57} &
  \multicolumn{1}{l|}{9.80} &
  11.74 \\ \hline
\multicolumn{1}{|l|}{\begin{tabular}[c]{@{}l@{}}RGB MobileViT-S +\\ DCT Effv2b0\end{tabular}} &
  Protocol I &
  \textbf{1.17} &
  \multicolumn{1}{l|}{\textbf{2.56}} &
  \multicolumn{1}{l|}{36.57} &
  \multicolumn{1}{l|}{5.99} &
  \multicolumn{1}{l|}{\textbf{7.73}} &
  \textbf{9.06} \\ \hline
\end{tabular}
\end{table}

\section{Conclusions} 
\label{Conclusions}

Minimum score fusion between spatial and frequency-domain features achieved the best performance. These findings suggest that integrating handcrafted frequency features with deep learning models enhances manipulation detection, demonstrating the effectiveness of this hybrid approach in improving robustness against various manipulation techniques.
\vspace{-0.3cm}
\section*{Acknowledgements}
This work was supported by Facephi, R\&D department and the German Federal Ministry of Education and Research and the Hessian Ministry of Higher Education, Research, Science and the Arts within their joint support of the National Research Center for Applied Cybersecurity ATHENE.
%
%
%
%
\printbibliography
\end{document}